\pdfoutput=1

\documentclass[11pt]{article}

\usepackage[]{naacl2021} 

\usepackage{times}
\usepackage{latexsym}

\usepackage[T1]{fontenc}

\usepackage[utf8]{inputenc}

\usepackage{microtype}

\usepackage{url}
\usepackage[T1]{fontenc}
\usepackage[draft]{todonotes}
\usepackage{amsmath,amsfonts,amssymb}
\usepackage{amsfonts}
\usepackage{graphicx}
\usepackage{multirow, tabu}
\usepackage[title]{appendix}
\usepackage{arydshln}
\usepackage[acronyms]{glossaries}
\usepackage{glossaries-prefix}
\usepackage{comment}
\usepackage{soul}
\usepackage{placeins}
\usepackage{enumitem}
\glsdisablehyper

\newcommand{\sectdot}[1]{Sec.~\ref{sec:#1}}

\newcommand{\fig}[1]{Figure~\ref{fig:#1}}

\newcommand{\tbl}[1]{Table~\ref{tab:#1}}

\newcommand{\twoeqndot}[2]{Eqns.~(\ref{eqn:#1}) and (\ref{eqn:#2})}

\newcommand{\ignore}[1]{}

\makeatletter
\DeclareRobustCommand\onedot{\futurelet\@let@token\@onedot}
\def\@onedot{\ifx\@let@token.\else.\null\fi\xspace}

\definecolor{MyDarkBlue}{rgb}{0,0.08,1}
\definecolor{MyDarkGreen}{rgb}{0.02,0.6,0.02}
\definecolor{MyDarkRed}{rgb}{0.8,0.02,0.02}
\definecolor{MyDarkOrange}{rgb}{0.40,0.2,0.02}
\definecolor{MyPurple}{RGB}{111,0,255}
\definecolor{MyRed}{rgb}{1.0,0.0,0.0}
\definecolor{MyGold}{rgb}{0.75,0.6,0.12}
\definecolor{MyDarkgray}{rgb}{0.66, 0.66, 0.66}

\newcommand\MyDarkBluesout{\bgroup\markoverwith{\textcolor{MyDarkBlue}{\rule[0.5ex]{2pt}{1.2pt}}}\ULon}

\newacronym{ann}{ANN}{artificial neural network}
\newacronym{abus}{ABUS}{agenda-based user simulator}
\newacronym{dst}{DST}{dialogue state tracking}
\newacronym{nlg}{NLG}{natural language generation}
\newacronym{rl}{RL}{reinforcement learning}
\newacronym{sl}{SL}{supervised learning}
\newacronym{pomdp}{POMDP}{partially observable markov decision process}
\newacronym[prefixfirst={a\ }, prefix={an\ }]{us}{US}{user simulator}
\newacronym{ds}{DS}{dialogue system}
\usepackage{xspace}
\usepackage{pdfpages}
\definecolor{green}{rgb}{0.0, 0.42, 0.24} 

\usepackage{xcolor,pifont}
\usepackage{stfloats}

\newcommand*\colourcheck[1]{%
  \expandafter\newcommand\csname #1check\endcsname{\textcolor{#1}{\ding{51}}}%
}
\colourcheck{blue}
\colourcheck{green}
\colourcheck{red}

\newcommand*\colourcross[1]{%
  \expandafter\newcommand\csname #1cross\endcsname{\textcolor{#1}{\ding{55}}}%
}
\colourcross{blue}
\colourcross{green}
\colourcross{red}

\title{CREAD: Combined Resolution of Ellipses and Anaphora in Dialogues}

\author{Bo-Hsiang Tseng$^{1}$\thanks{$^{*}$Work done while the first author was an intern at Apple.}, Shruti Bhargava${}^2$, Jiarui Lu${}^2$ \\ \textbf{Joel Ruben Antony Moniz${}^2$, Dhivya Piraviperumal${}^2$, Lin Li${}^2$, Hong Yu${}^2$} \\
${}^1$Engineering Department, University of Cambridge, UK \\
${}^2$Apple \\
\texttt{bht26@cam.ac.uk} \\
\texttt{\{shruti\_bhargava, jiarui\_lu,}\\ 
\texttt{joelrubenantony\_moniz, dhivyaprp, lli9, hong\_yu\}@apple.com}\\
}
\begin{document}
\maketitle

\begin{abstract}
Anaphora and ellipses are two common phenomena in dialogues.
Without resolving referring expressions and information omission, dialogue systems may fail to generate consistent and coherent responses.
Traditionally, anaphora is resolved by coreference resolution and ellipses by query rewrite.
In this work, we propose a novel joint learning framework of modeling coreference resolution and query rewriting for complex, multi-turn dialogue understanding.
Given an ongoing dialogue between a user and a dialogue assistant, for the user query, our joint learning model first predicts coreference links between the query and the dialogue context, and then generates a self-contained rewritten user query.
To evaluate our model, we annotate a dialogue based coreference resolution dataset, MuDoCo, with rewritten queries.
Results show that the performance of query rewrite can be substantially boosted (+2.3\% F1) with the aid of coreference modeling.
Furthermore, our joint model outperforms the state-of-the-art coreference resolution model (+2\% F1) on this dataset.

\end{abstract}
\section{Introduction}

\begin{figure}[t]
    \centering
    \includegraphics[width=1\linewidth]{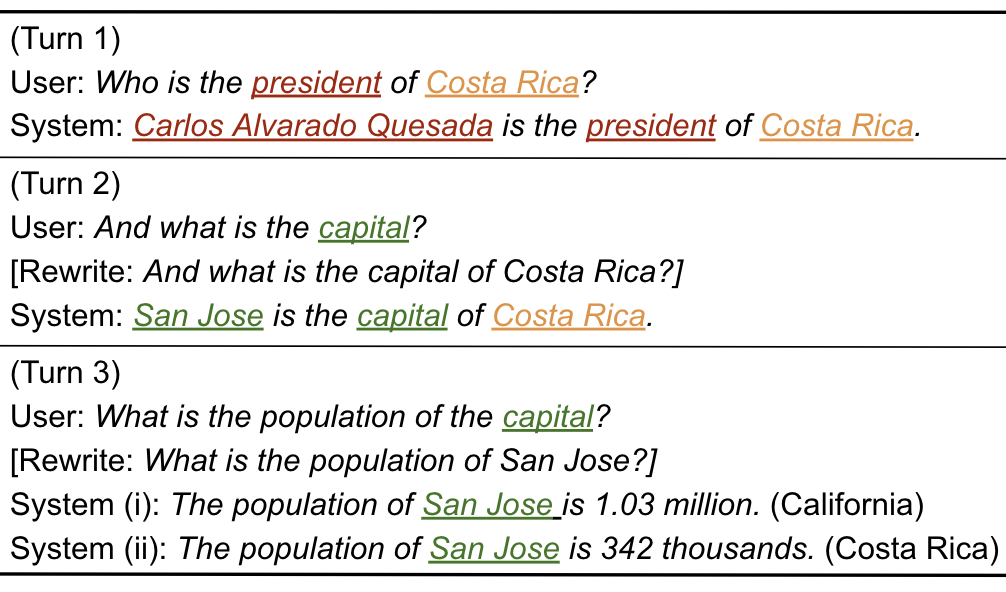}
    \caption{An example of a question-answering dialogue where coreference and ellipsis happen in user query, and the corresponding query rewrite annotation. References to the same entity are highlighted in the same color, and can be resolved by coreference resolution modeling. The two system responses in Turn 3 indicate two possible interpretations of the city \textit{San Jose} by the system. }
    \label{fig:example_intro}
\end{figure}

In recent years, dialogue systems have attracted growing interest,
and been applied to various scenarios, ranging from chatbots to task-oriented dialogues to question answering. Despite rapid progress in dialogue systems, several difficulties remain in the understanding of complex, multi-turn dialogues. Two major problems are anaphora resolution \cite{clark2016deep,clark2016improving} and ellipsis \cite{kumar-joshi-2016-non} in follow-up turns.
Take the dialogue in \fig{example_intro} as an example: ellipsis happens in \texttt{user turn 2} where the user is asking for the capital of \textit{``Costa Rica''} without explicitly mentioning the country again; coreference happens in \texttt{user turn 3} where \textit{``the capital''} refers to \textit{``San Jose''}.
Without resolving the anaphoric reference and the ellipsis, dialogue systems may fail to generate coherent responses.


Query rewrite \cite{quan2019gecor} is an approach that converts a context-dependent user query into a self-contained utterance so that it can be understood and executed independent of previous dialogue context. This technique can solve many cases where coreference or ellipsis happens.
For instance, ``\textit{the capital}'' in \texttt{user turn 3} is changed to \textit{``San Jose''} in the rewrite. Furthermore, the ellipsis of the country name \textit{``Costa Rica''} in \texttt{user turn 2} can be revealed through rewriting.
The rewritten utterance improves multi-turn dialogue understanding \cite{yang-etal-2019-end-end} by reducing dependency on the previous turns.

Although query rewrite implicitly resolves coreference resolution, there is information not contained in a rewrite.
First, it does not provide a distinct coreference link between mentions across dialogue turns as in the classic coreference resolution task.
This is particularly disadvantageous when there is entity ambiguity in the rewritten sentence.
For example, in Figure \ref{fig:example_intro}, since \textit{``San Jose''} in \texttt{Rewrite turn 3} can be either San Jose in Costa Rica or San Jose in California, it is likely that the system ends up with an incorrect response by generating System (i) instead of System (ii) due to the wrong interpretation of San Jose.
Second, mention detection, an essential step in coreference resolution \cite{peng-etal-2015-joint}, is not involved in query rewrite.
By knowing which span in an utterance is a mention, downstream systems like named entity recognition and intent understanding can perform better \cite{bikel2009entity}.
Third, if coreference links to dialogue context are available, downstream systems can skip entity linking, which is time-consuming and may introduce noise.

To resolve the above issues, we propose a novel joint learning framework that incorporates the benefits of reference resolution into the query rewrite task. To the best of our knowledge, there does not exist, at the time of writing, an English conversational dataset that couples annotations of both query rewrite and coreference resolution (as links or clusters).
This motivates us to collect annotations for query rewrite on a recent dialogue dataset - MuDoCo \cite{martin2020mudoco}, which already has coreference links between user query and dialogue context.
Compared to existing query rewrite datasets \cite{quan2019gecor,anantha2020open}, rewriting in MuDoCo is much more challenging since it involves reasoning over multiple turns and spans multiple domains.

We design a joint learning model adopting the GPT-2 \cite{radford2019language} architecture that learns both query rewrite and coreference resolution.
Given an ongoing dialogue, our model first predicts the coreference links, if any, between the latest user query and the dialogue context. 
Then it generates the rewritten query by drawing upon the coreference results.
Our experiments show that query rewrite performance can be substantially boosted with the aid of coreference training. In addition, our model outperforms strong baselines for the two individual tasks. Since both the tasks fundamentally solve reference resolution, the joint training facilitates knowledge sharing.

Our contributions can be summarized as follows:
\begin{itemize} 
    \item We present a novel joint learning framework of modeling coreference resolution and query rewrite for multi-turn dialogues.
    \item Our annotations of query rewrite augment the MuDoCo dataset with query rewrite labels. To the best of our knowledge, our augmented MuDoCo is the first English dialogue dataset with both coreference resolution and query rewrite annotations.
    \item We propose a novel GPT-2 based model to tackle the two target tasks, and show that joint training with coreference resolution helps in improving the quality of the query rewrites.
    
\end{itemize}
The augmented dataset with our annotations along with the modeling source code are available at \url{https://github.com/apple/ml-cread}.


\section{Related Work}

\paragraph{Query Rewrite}
The most relevant line of research is the adoption of query rewrite in dialogues to tackle anaphora or ellipses.
Many prior works employ an LSTM-based seq-to-seq model, which takes the dialogue context and user query as input, and generates the rewritten query.
\citet{quan2019gecor} use the pointer-generator model \cite{see2017get} to rewrite the user query on restaurant-domain task-oriented dialogues.
By comparison, query rewrite in MuDoCo dataset is more challenging as it covers 6 domains and the rewriting patterns are more complex and diverse than in the CamRest676 dataset \cite{wen2017network}.
\citet{rastogi2019scaling} introduce an auxiliary objective of copying entity tokens from the delexicalized utterances to augment the learning of pointer network.
In \citet{su2019improving}, two separate attention distributions are learned for the dialogue context and the user query respectively with a control gate.
This modified copy mechanism shows improvements over the standard pointer-generator on both LSTM-based models and transformer-based models \cite{vaswani2017attention}.
Note that in the dataset used in their work, the dialogue context has only 2 utterances; MuDoCo, in contrast, has up to 8 utterances, making it much more challenging for query rewrite.

\paragraph{Coreference Resolution}
Research on document-based coreference resolution has a long history (a detailed survey can be found in \citet{ng2010supervised}).
Various approaches have been proposed, ranging from learning mention-pair classifiers \cite{ng2002identifying,bengtson2008understanding}, latent structured-based models \cite{fernandes2012latent,bjorkelund2014learning,martschat2015latent} to the more recent neural pipeline based systems that rely on syntactic parsers \cite{raghunathan2010multi} and clustering algorithms \cite{clark2016deep,clark2016improving}.
The first neural end-to-end coreference resolution model was proposed in \citet{lee2017end} and achieved better results without external resources. An improved version was proposed in \citet{lee2018higher}, which considers higher-order structures by iteratively refining span representations. Recently, powerful pre-trained models have been used to extract representations for these end-to-end models using BERT \cite{joshi2019bert} or SpanBERT \cite{joshi2020spanbert}. \citet{wu-etal-2020-corefqa} approach the problem in a question answering framework. For each detected mention candidate, the sentence it resides in serves as the query and is used to predict the referent in the passage.
Different from these works, we focus on coreference resolution in dialogues with the following main distinctions:
1) the speaker information in dialogues is clear; 2) less descriptive content may cause the pronoun mention to be more ambiguous; and 3) coreference resolution is conducted only between the latest user query and the previous dialogue context
--- unlike in document-based coreference resolution where a model can look ahead for the resolution, future turns are not available to a dialogue agent.
We encourage the reader to refer to \citet{martin2020mudoco} for more details.


\paragraph{Joint Learning}
In contrast to prior works that focus solely on either query rewrite or coreference resolution, we present a novel joint learning approach to tackle both the tasks using one single model. We hope that this work serves as a first step towards this new, challenging and practical problem for dialogue understanding.



\section{Dataset and Task} \label{sec:task}

\begin{figure}[t]
    \centering
    \includegraphics[width=1\linewidth]{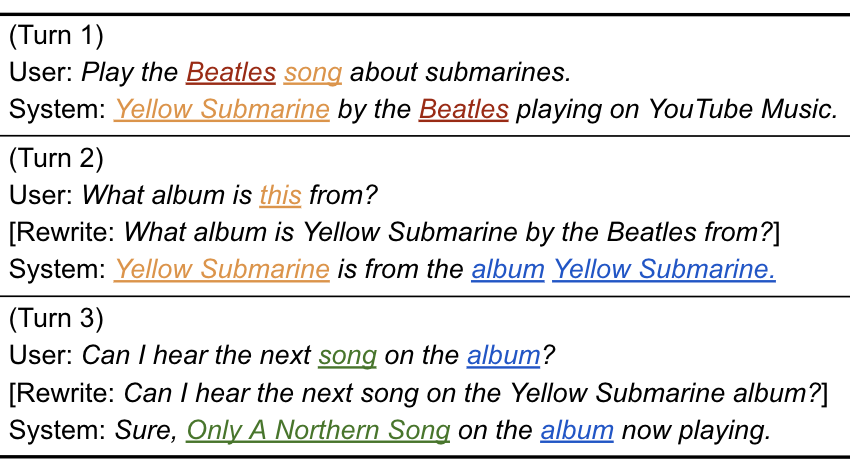}
    \caption{An example from the MuDoCo dataset in the music domain with our query rewrite annotation. Word spans in the same color belong to the same mention cluster.}
    \label{fig:example_mudoco}
\end{figure}

The MuDoCo dataset contains 7.5k task-oriented multi-turn dialogues across 6 domains.
A dialogue has an average of 2.6 turns and a maximum of 5 (a turn includes a user query and a system response).
\fig{example_mudoco} shows an example. For each partial dialogue, the coreference links, if existing, are annotated between the latest user query and its dialogue context. For example, when we consider the partial dialogue preceding up to \texttt{user turn 2}, there is a coreference link between the anaphora ``\textit{this}'' in \texttt{user turn 2} and the antecedent ``\textit{song}'' in \texttt{user turn 1}.
When an anaphora has multiple antecedents in the context, e.g., ``\textit{song}'' in \texttt{user turn 1} and ``\textit{Yellow Submarine}'' in \texttt{system turn 1}, only one of them is annotated as its referent in the coreference link.


On top of the existing coreference labels, we annotate the rewrite for each utterance.
The goal is to rewrite the query into a self-contained query independent of the dialogue context.
30 annotators are recruited for the data collection. Each of them is shown a partial dialogue, and is asked (1) to decide if the query needs to be rewritten due to coreference or ellipsis; and (2) to provide the rewritten query, when rewriting is required.
We notice that there can be various ways of rewriting an utterance. For example, some annotators might include every detail of the rewritten entity, while others might choose a precise term; some might paraphrase the rewritten utterance, while others keep the same expression. To ensure data consistency and high annotation quality, we designed a comprehensive guideline for the annotators to follow and undertook a two-stage collection process: 1) we organized two training sessions with annotators. In each session, 50 representative examples were selected and assigned to each annotator. An author inspected these training results individually and provided feedback to the annotators. 2) 5\% of the grading results were manually evaluated by an author for quality assurance.
Detailed annotation guidelines can be found in the Appendix.

The joint learning task requires the machine to predict both coreference links and the rewritten query for the latest user query given an ongoing dialogue.
The outputs of the two individual tasks complement each other and provide more comprehensive information for dialogue understanding.
For instance, the ``\textit{Yellow Submarine}'' in \fig{example_mudoco} can be either a song name or an album name. Explicit coreference resolution helps to disambiguate between various possibilities by linking entities to previously resolved ones. More importantly, the supervision of coreference resolution can be beneficial to rewriting the anaphora to its antecedent.


\section{Modeling}
Our proposed model for jointly learning coreference resolution and query rewrite is designed based on the GPT-2 architecture, presented in \fig{model}. 
The input to the model is the concatenation of the dialogue context and the latest user query, where special tokens are used to separate utterances and indicate speaker information.
Passing through the standard decoder layers, the hidden state $h^{l}_{t} \in \mathbb{R}^{d}$ and attention score $a^{l,j}_{t} \in \mathbb{R}^{T}$ at each position of the input sequence are calculated, where $l$, $j$ and $t$ denote the index of the decoder layer, that of the attention head, and the input token position respectively; $d$ and $T$ denote the embedding size and the length of the input sequence respectively.
Inspired by the end-to-end coreference resolution model \cite{lee2017end}, our model first predicts mentions in the user query and grounds them to their corresponding referent in the dialogue context using attention heads.
The model then generates the rewritten query conditioned on the resolved coreference links.
The prediction process has four main steps, described in detail below:

\begin{figure*}[t]
    \centering
    \includegraphics[width=0.9\linewidth]{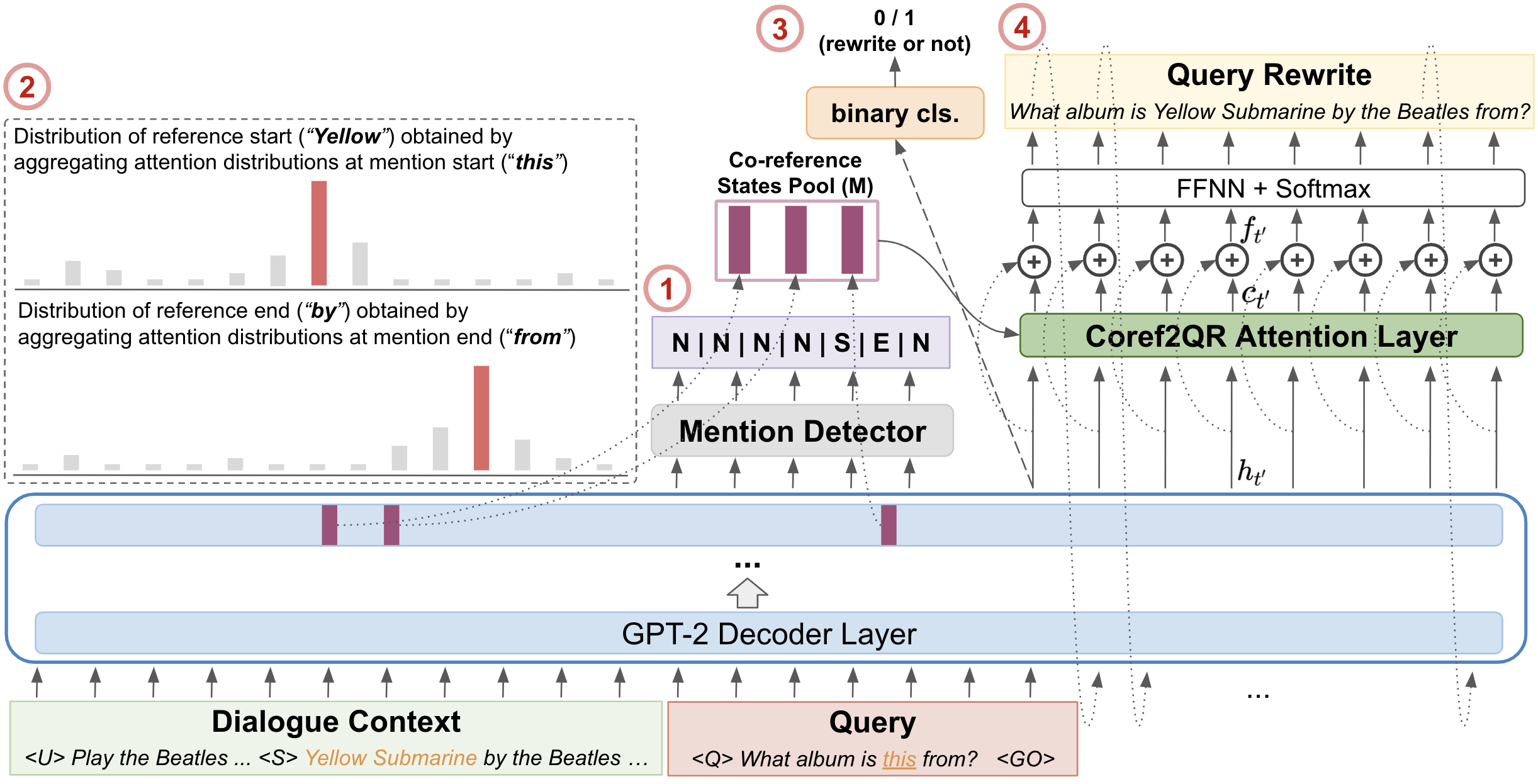}
    \caption{The proposed model for joint learning of coreference resolution and query rewrite, designed using the GPT-2 architecture. Given a dialogue context and a user query, the model first detects the mentions in the query (Step 1); resolves the corresponding reference spans (Step 2); predicts whether the query needs a rewrite or not (Step 3); and, if the model decides to rewrite, generates the rewritten query (Step 4). In this example dialogue, there is a coreference link existing between the mention ``\textit{this}'' and its referent ``\textit{Yellow Submarine}''.}
    \label{fig:model}
\end{figure*}

\paragraph{Step 1: Mention Detection}
First, the model detects any possible referring expressions in the user query.
Here we use the term \texttt{mention} to include all those expressions that require reference resolution (e.g., pronouns or partial entity names).
We formulate mention detection as a sequence labeling problem: each token in a query is labelled as one of three classes $\{S,E,N\}$, referring to Start of mention, End of mention and None respectively.
This sequence tagger in the mention detector, parameterized by a feed-forward network, takes the hidden states of the query from the last decoder layer as input, and predicts the sequence of class labels.
Then the mention spans in the query can be determined by a pair of mention start $S$ and end $E$ tags.
For instance, in \fig{model} the label of position ``\textit{this}'' is class $S$ and that of ``\textit{from}'' is class $E$, while the rest of the positions in the query are labelled as class $N$.
We use $m_{S}$ and $m_{E}$ to respectively denote the start and end position index of a predicted mention $m$.

\paragraph{Step 2: Reference Resolution}
For each detected mention $m$, the model resolves it to the antecedent (or referent) in the dialogue context by predicting the span boundaries: the position index of the referent start $r_{S}$ and end $r_{E}$.
Essentially, the distributions of the boundaries ($r_{S}$ and $r_{E}$) are learned by supervising multiple attention heads associated with the target mention $m$.
In other words, the attention distribution $a_{m_{S}}$ (the attention score of each position associated with the mention start $m_{S}$) is supervised to focus on the referent start $r_{S}$.
Similarly, attention scores $a_{m_{E}}$ associated with the  mention end $m_{E}$ are used to learn the boundary of referent end $r_{E}$.
Concretely:

\begin{equation} \label{eqn:ref_dis}
\begin{aligned}
    q_{r_{S}} = \frac{1}{L'J'} \sum^{L'}_{l} \sum^{J'}_{j} a^{l,j}_{m_{S}}\,, \\
    q_{r_{E}} = \frac{1}{L'J'} \sum^{L'}_{l} \sum^{J'}_{j} a^{l,j}_{m_{E}}\,,
\end{aligned}
\end{equation}
where $q_{r_{S}}$ and $q_{r_{E}}$ are the probability distributions that a given token represents the referent start $r_{S}$ and end $r_{E}$ respectively; $L'$ and $J'$ are the specified number of the involved decoder layers and attention heads.
We then take the argmax of these boundary distributions to resolve the referent $r$. 
Our design of reference resolution effectively leverages the powerful attention mechanism in GPT-2 without adding any extra components for reference resolution.

\paragraph{Step 3: Binary Rewriting Classification}
The model completes the coreference resolution in steps 1 and 2, after which it starts producing the rewritten query.
Unlike existing query rewrite systems that directly generate the rewrite given the input, our model first predicts whether the incoming query requires to be rewritten using a binary classifier.
As shown in \fig{model}, the classifier, a two-layer feed-forward network followed by a softmax layer, takes as input the hidden state of the first decoding step and predicts a vector with two entries representing the rewrite and no-rewrite classes.
Only when the binary prediction is true, i.e., the classifier predicts the class indicating that a rewrite is required, does the model enter Step 4 to generate the rewritten query; otherwise, the input query will be directly copied as the output.
We show that a well-learned binary classifier with 93\% accuracy functions as a filter that helps the model not only minimize the risk of incorrectly rewriting already self-contained queries, but also allows the rest of the generation process to solely focus on how to rewrite incomplete queries during training.

\paragraph{Step 4: Query Rewrite Generation}
In this final step, the model runs the generation step based on its binary decision of whether or not to rewrite. Unlike the standard language modeling setup in GPT-2, where the output sequence is generated directly from the last hidden states, we design the \texttt{Coref2QR} attention layer that allows information gained during coreference resolution to effectively assist in the query rewrite generation.

First, all relevant hidden states of mentions and referents predicted in Steps 1 and 2 are assembled to form a memory pool $M$. Note that it is possible for an example to have more than one coreference link. At each time step $t'$ during the rewrite generation, the Coref2QR attention layer, operating as the standard multi-head attention mechanism, takes $h_{t'}$ as query to attend over the coreference related states $M$ by treating them as keys and values. The resulting attention head $c_{t'}$ is summed with $h_{t'}$ to obtain the feature $f_{t'}$ before the final output token classifier.
This design improves information flow between the two tasks, enabling the model to directly utilize information regarding previously resolved coreferents during rewrite generation.

The Coref2QR attention can be applied to any arbitrary decoder layer to facilitate the deeper interaction between rewrite and coreference resolution in the model. Formally, at each decoder layer $l$, the memory pool $M^{l}$ stores coreference related states produced at layer $l$. At the generation step $t'$, the Coref2QR layer takes $h^{l}_{t'}$ as query to attend over $M^{l}$ to obtain $c^{l}_{t'}$. The final feature $f_{t'}$ before the output token classifier is then obtained by $f_{t'} = h^{L}_{t'} + \frac{1}{L} \sum_{l} c^{l}_{t'}$. For simplicity, in \fig{model} we only illustrate the Coref2QR attention for the last decoder layer.
Our results and analysis show that this Coref2QR attention design benefits the quality of query rewrite, especially in rewriting an anaphora into its antecedent.

\paragraph{Optimization}
During training, an input sequence with length $T$ is formed by the concatenation of the dialogue context, the user query and the target query rewrite. Four objectives, corresponding to each step in the model, are used for training.
For mention detection, the objective is the cross-entropy between the predicted sequence of mention class $p^{M}$ and its ground-truth sequence $y^{M}$:
\begin{equation} \label{eqn:men}
    L^{M} = \sum^{q_{E}}_{t=q_{S}} -\log ((\mathbf{y}^{M}_{t})^{\top} \mathbf{p}^{M}_{t})\,,
\end{equation}
where $q_{S}$ and $q_{E}$ denote the start and end index of the query respectively. $\top$ is the transpose operation.

For each coreference link $n$, the loss is calculated using the cross-entropy between the predicted distributions of the antecedent boundaries $q^{n}$ and the corresponding ground-truth $y^{R_{n}}$. The final loss for reference resolution is the sum of losses from the existing coreference links:
\begin{equation} \label{eqn:ref}
    L^{R} = \sum^{N}_{n=1} -\log ((\mathbf{y}^{R_{n}}_{r_{S}})^{\top} \mathbf{q}^{n}_{r_{S}})
    -\log ((\mathbf{y}^{R_{n}}_{r_{E}})^{\top} \mathbf{q}^{n}_{r_{E}})\,,
\end{equation}
where $N$ is the number of coreference links in an example. $\mathbf{q}^{n}_{r_{S}}$ and $\mathbf{q}^{n}_{r_{E}}$ represent the predicted distributions of reference start $r_{S}$ and reference end $r_{E}$ respectively. When an example does not contain a coreference link, $L^{R}$ would be $0$.

For query rewrite, the binary classification loss is the two-class cross entropy between the prediction $p^{B}$ and the binary rewriting label $y^{B}$:
\begin{equation} \label{eqn:bi}
    L^{B} = -\log ((\mathbf{y}^{B})^{\top} \mathbf{p}^{B})
\end{equation}
For generation, as in the standard language modeling task, we use cross-entropy between the predicted sequence $p^{Q}$ and its ground-truth sequence $y^{Q}$: 
\begin{equation} \label{eqn:qr}
    L^{Q} = \sum^{T}_{t'=q_{E}+1} -\log ((\mathbf{y}^{Q}_{t'})^{\top} \mathbf{p}^{Q}_{t'})
\end{equation}
where $t'$ is the time step in the word sequence of query rewrite. Note that $L^{Q}$ is 0 for examples that do not need rewrite. 
The final loss is the sum of all these losses:
\begin{equation}
    L = L^{M} + L^{R} + L^{B} + L^{Q}.
\end{equation}



\section{Experiments}
\paragraph{Dataset}
As discussed in \sectdot{task}, we conduct all experiments on the MuDoCo dataset and follow the provided data split\footnote{\url{https://github.com/facebookresearch/mudoco}}. Data from 6 domains are aggregated to form train/dev/test sets with 16k, 1.9k and 1.9k examples respectively. Each example contains the dialogue context, the latest user query, and the corresponding coreference resolution and query rewrite annotations. Statistics for each domain are provided in \tbl{stats}. 
Out of all examples that are not the first turn, 64.2\% of them contain coreference links and 43.7\% of them require query rewrite.
This makes the task more challenging, as the model also needs to learn when not to rewrite a query and when to predict no coreference links.
Note that not every coreference link requires rewriting, as in the MuDoCo dataset there are coreference annotations where the mention has the exact same word span as its referent.

\paragraph{Setup}
The GPT-2 decoder layers and word classification layer in our model are initialized with the pre-trained weights from the GPT-2 small model. We fine-tune the model using Adam \cite{kingma2014adam} optimizer with learning rate 5e-05 and batch size 15. The criterion for early stopping is the averaged performance of coreference resolution and query rewrite on the development set.
Results are obtained as the average of 5 runs.


\begin{table}[t]
\centering
\resizebox{0.9\linewidth}{!}{%
\begin{tabu}{lccc}
\tabucline [1pt]{1}
\textbf{Domain}    & \textbf{Total} & \textbf{Coref.} & \textbf{Rewrite} \\ \hline
Calling   & 10.7k & 4.0k (60.5\%)   & 2.2k (33.7\%)    \\
Messaging & 3.9k  & 1.7k (69.0\%)   & 1.0k (41.2\%)    \\
Music     & 2.8k  & 1.4k (77.7\%)   & 1.4k (76.7\%)   \\
News      & 387   & 156  (66.4\%)   & 189 (76.6\%)     \\
Reminders & 1.7k  & 632  (56.4\%)   & 357 (31.9\%)     \\
Weather   & 254   & 38  (28.6\%)   & 102 (76.6\%)     \\ \hline
All       & 19.8k & 8.0k (64.2\%)   & 5.3k (43.7\%)   \\
\tabucline [1pt]{1}
\end{tabu}%
}
\caption{Total number of examples across six domains in the MuDoCo dataset, number of examples requiring coreference resolution (Coref.) and those that need query rewrite (Rewrite). Percentages are calculated across all follow-up turns (i.e., excluding the first turn).
}
\label{tab:stats}
\end{table}

\subsection{Query Rewrite}
\paragraph{Evaluation Metrics}
The standard BLEU-4 \cite{papineni2002bleu} between the generated and the target sentences are reported. In addition, to highlight the quality of the rewritten parts in generated sentences, following the post-processing in \citet{quan2019gecor}, we measure an F1 score calculated by comparing machine-generated words with ground truth words for only the ellipsis / co-reference part of user utterances.
We also report the percentage of all referents in ground-truth coreference links that were successfully generated in the query rewrite, denoted as reference match (RM).
The RM ratio explicitly reflects the quality of coreference resolution in the generated rewritten query.

\paragraph{Baselines}
The standard seq-to-seq model with attention (seq2seq) and its pointer network (PN; \citet{vinyals2015pointer}) and pointer-generator network (PG; \citet{see2017get}) variants are implemented as baselines. 
The concatenation of the dialogue context and the query are fed as input, and the output is the target rewrite. The size of the hidden states is 300 and word vectors are initialized with GloVe embeddings \cite{pennington2014glove}.

\begin{table}[t]
\centering
\resizebox{\linewidth}{!}{%
\begin{tabu}{lcccccc}
\tabucline [1pt]{1}
\textbf{Model}         & Prec. & Rec. & F1   & BLEU & RM \\ \hline
seq2seq model          & 38.3           & 29.6          & 33.4          & 81.0    & 54.7                \\
+ pn \cite{vinyals2015pointer}      & 42.4           & 34.1          & 37.6          & 86.0      & 61.2            \\
+ pg \cite{see2017get}    & 41.4           & 39.5          & 40.4          & 86.4      & 63.2           \\ \hdashline
Our QR-only model      & 58.9           & 57.1          & 57.9          & 89.8      & 78.7               \\
Our joint model        & \textbf{61.0}  & \textbf{59.5} & \textbf{60.2} & \textbf{90.2}  &\textbf{82.0} \\
\tabucline [1pt]{1}
\end{tabu}%
}
\caption{Query rewrite results in F1 and BLEU. QR-only model is our model variant trained using only objectives of query rewrite.}
\label{tab:qr_result}
\end{table}

\paragraph{Results}
\tbl{qr_result} shows the query rewrite results. The low F1 score and high BLEU score is because of filtering out the non-rewritten repeated tokens in post-processing when calculating F1. This allows us to better evaluate the quality of rewritten parts and to better differentiate between good and bad generation in our task.
We find that our joint model substantially outperforms all LSTM-based seq-to-seq models on all metrics.
Although the pointer-generator in LSTMs can effectively copy words from the input to its generation, the powerful transformer architecture with pre-trained weights allows better learning of rewriting patterns.

To fairly investigate the impact of coreference modeling
on the generation of query rewrite, we train a variant of our model using only the query rewrite objectives (\twoeqndot{bi}{qr}), denoted as QR-only model. We can see that without coreference resolution, the F1 score drops from 60.2 to 57.9 and the reference match drops from 82.0 to 78.7. 
This illustrates the improved ability of the joint model to rewrite anaphoric expressions, since the model can leverage its coreference resolution predictions to generate more accurate query rewrites.
We present a detailed case study with model predictions in \sectdot{case_study}.



\begin{table*}[b]
\centering
\resizebox{0.9\linewidth}{!}{%
\begin{tabu}{lcccccccccc}
\tabucline [1pt]{1}
                     & \multicolumn{3}{c}{MUC}      & \multicolumn{3}{c}{$\textup{B}^{\textup{3}}$}       & \multicolumn{3}{c}{$\textup{CEAF}_{\phi_{4}}$} & \textbf{}        \\
                     & P & R & F1 & P & R & F1 & P  & R  & F1 & Avg. F1 \\ \hline
c2f-coref + BERT \cite{joshi2019bert}     & 72.2       & 66.7       & 69.3        & 74.5       & 67.9       & 71.0        & 77.7        & 72.6        & 75.1        & 71.8             \\
c2f-coref + SpanBERT \cite{joshi2020spanbert} & 71.7       & \textbf{71.4}       & 71.5        & 73.5       & \textbf{72.5}       & 73.0        & 77.8        & 74.9        & 76.3        & 73.6             \\ \hdashline

Our coref-only model & \textbf{78.8} & 69.4 & \textbf{73.8} & \textbf{79.6} & 71.3 & \textbf{75.2} & 80.7 & 75.1 & 77.8 & \textbf{75.6} \\

Our joint model      & 78.3       & 69.4       & 73.6        & 79.5       & 71.2       & 75.1        & \textbf{81.1}        & \textbf{75.1}        & \textbf{78.0}        & \textbf{75.6}            \\
\tabucline [1pt]{1}
\end{tabu}%
}
\caption{Coreference resolution results. ``Our coref-only model'' is our model variant trained only using the objectives of coreference resolution.}
\label{tab:coref_result}
\end{table*}

\subsection{Coreference Resolution}
\paragraph{Evaluation Metrics}
The MUC, $\textup{B}^{\textup{3}}$, and $\textup{CEAF}_{\phi_{4}}$ metrics that are widely-used in coreference resolution task are reported. Note that these metrics are calculated based on coreference clusters and we only have ground-truth annotations for coreference links between mentions and referents. To align the links and clusters, during evaluation we post-process both the ground-truth and the model predictions.
All the word spans that are identical to the referent in the dialogue context are combined into a cluster so that a link between a mention and a referent can be transformed into a cluster for the standard coreference resolution evaluation.


\paragraph{Baselines}
To the best of our knowledge, there is no suitable coreference resolution model that is proposed in the same setup for dialogues\footnote{The baseline in \citet{martin2020mudoco} is not compared for two reasons: 1) their setups in training/evaluation is different than ours in many ways, e.g., they only consider finished dialogues; 2) their source code is not released.}.
We therefore experiment with the state-of-the-art models of document-based coreference resolution, including the end-to-end model \cite{lee2017end,lee2018higher} using BERT \cite{joshi2019bert} or SpanBERT \cite{joshi2020spanbert}\footnote{\url{https://github.com/mandarjoshi90/coref}}. Note that these models can only serve as a reference since they are not specifically designed for dialogue-based tasks. Since they require coreference clusters for training, coreference clusters are built from annotated links as in the post-processing step done for evaluation.

\paragraph{Results}
As seen in \tbl{coref_result}, SpanBERT obtains better results than BERT, which is consistent with the findings in \citet{joshi2020spanbert}. This is mainly because SpanBERT is better at capturing span information, which facilitates tasks such as coreference resolution where reasoning about relationships between spans is required.
In comparison, our joint learning model achieves competitive and even slightly better results. This indicates that the design of our model leveraging attention heads inside GPT-2 is effective at predicting coreference links in dialogues.
To test if the supervision of query rewrite affects the optimization of coreference resolution in joint learning, we train a model variant using only the objectives for coreference resolution (\twoeqndot{men}{ref}), denoted as \texttt{coref-only} model. It is observed that the results of the coref-only model are very close to that of the joint model, showing that the addition of coreference resolution in joint learning is beneficial to query rewrite without sacrificing the performance of the former.
%

\begin{table}[t]
\centering
\resizebox{\linewidth}{!}{%
\begin{tabu}{lccccc}
\tabucline [1pt]{1}
                        & Prec. & Rec. & F1   & BLEU & RM   \\ \hline
complete model & \textbf{61.0} & \textbf{59.5} & \textbf{60.2} & \textbf{90.2} & \textbf{82.0} \\
- coref2qr attention    & 55.5  & 59.3 & 57.3 & 89.3 & 80.6 \\
- coref. modeling       & 58.9  & 57.1 & 57.9 & 89.8 & 78.7 \\
- binary head           & 54.6  & 54.4 & 54.3 & 88.9 & 78.9 \\
\tabucline [1pt]{1}
\end{tabu}%
}
\caption{Ablation study of our joint learning model on query rewrite performance.}
\label{tab:ablation_qr}
\end{table}

\begin{table}[t]
\centering
\resizebox{\linewidth}{!}{%
\begin{tabu}{l|cc|cc|cc|cc}
\tabucline [1pt]{1}
\multirow{2}{*}{Model} & \multicolumn{2}{c|}{Calling}  & \multicolumn{2}{c|}{Messaging} & \multicolumn{2}{c|}{Music}    & \multicolumn{2}{c}{All}       \\
        & coref. & elp. & coref. & elp.          & coref. & elp. & coref. & elp. \\ \hline
seq2seq+pg & 56.0   & 36.2 & 63.6   & 36.2          & 45.5   & 38.2 & 52.0   & 34.5 \\
QR-only & 75.4   & 51.4 & 77.6   & \textbf{66.5} & 59.8   & 45.5 & 69.0   & 49.3 \\
Joint                  & \textbf{78.3} & \textbf{52.0} & \textbf{81.3}      & 64.3      & \textbf{63.1} & \textbf{51.1} & \textbf{72.1} & \textbf{50.9} \\
\tabucline [1pt]{1}
\end{tabu}%
}
\caption{Query rewrite performance (F1) over three main domains and All test set with respect to two types of rewriting: coreference (coref.) and ellipses (elp.).}
\label{tab:analysis}
\end{table}

\definecolor{MyDarkBlue}{rgb}{0,0,0.75}
\definecolor{MyDarkGreen}{rgb}{0,0.5,0}
\definecolor{MyDarkRed}{rgb}{0.8,0.02,0.02}
\definecolor{MyDarkOrange}{rgb}{0.40,0.2,0.02}
\definecolor{MyOrange}{rgb}{0.85,0.4,0}
\definecolor{MyPurple}{RGB}{111,0,255}
\definecolor{MyRed}{rgb}{1.0,0.0,0.0}
\definecolor{MyGold}{rgb}{0.75,0.6,0.12}
\definecolor{MyDarkgray}{rgb}{0.66, 0.66, 0.66}
\definecolor{MyYellow}{rgb}{0.7,0.7,0}

\begin{table*}[hb]
\centering
\resizebox{\linewidth}{!}{%
\begin{tabu}{l|l|l}
\tabucline [1pt]{1}
\begin{tabular}[c]{@{}c@{}}Dialogue\\Context\end{tabular} &
  \begin{tabular}[c]{@{}l@{}}
  usr: When was \textcolor{MyDarkGreen}{Talking to the Moon} by Bruno Mars released?\\ 
  sys: On April 12, 2011.\\ 
  usr: Who produced the song?\\ 
  sys: The Smeezingtons and Bhasker.
  \end{tabular} &
  \begin{tabular}[c]{@{}l@{}}
  usr: I want to send a message.\\ 
  sys: Who do you want to send a message to?\\ 
  usr: To Ariana.\\ 
  sys: Ariana Smith or \textcolor{MyDarkGreen}{Ariana Taylor}?
  \end{tabular} \\ \hline
User Query   & Could you play the \textcolor{MyOrange}{song} for me? & The second \textcolor{MyOrange}{one}. \\ \hline
Rewrite Label & Could you play \textbf{Talking to the Moon} for me?  & \textbf{Ariana Taylor}.                       \\ \tabucline [1pt]{1}
seq2seq+pg  & Could you play \textbf{the moon} for me?  \redcross   & \textbf{Ariana Smith}.  \redcross                     \\ \hline
QR-only  & Could you play the song for me?  \redcross    & \textbf{Ariana Smith}.  \redcross                    \\ \hline
Joint    & Could you play \textbf{Talking to the Moon} for me? (\textcolor{MyOrange}{song} -> \textcolor{MyDarkGreen}{Talking to the Moon}) \greencheck & \textbf{Ariana Taylor}. (\textcolor{MyOrange}{one} -> \textcolor{MyDarkGreen}{Ariana Taylor}) \greencheck \\
\tabucline [1pt]{1}
\end{tabu}%
}
\caption{Two coreference examples from test set with rewrites generated by three models: 1. seq-to-seq model with pointer-generator (seq2seq+pg); 2. Our QR-only model; 3. Our joint learning model. The rewritten parts are highlighted in bold. The coreference links predicted by 3. are presented as (mention -> antecedent).}
\label{tab:example_coref}
\end{table*}

\subsection{Ablation Study}
Here, we investigate how the different components in our joint model contribute to the performance of query rewrite. We remove one component at a time and examine the performance of query rewrite. As shown in \tbl{ablation_qr}, without the designed coref2qr attention layer, the performance degrades with a drop of 2.9\% F1 and 1.4\% RM rate. By further removing the supervision of coreference modeling from our joint learning model, the model is solely optimized towards the objectives of query rewrite and produces worse results compared to the complete model. These results indicate that through joint learning, the model's ability of generating the rewritten query improves, including its ability to rewrite the anaphora with its antecedent, by leveraging the information from coreference resolution modeling. In addition, the binary head plays an essential role in our model. The accuracy of this binary classifier is 93.9\%. Without the binary head, the performance drop can be up to 5.9\% F1 (60.2 -> 54.3). This shows that with the binary classification, the model is able to focus on rewriting the input query without worrying about whether to rewrite or not.

\subsection{Analysis}
In this section we analyze query rewrite performance on two different types of rewriting, coreference (coref.) and ellipses (elp.). The F1 score over three main domains and all test sets are reported in \tbl{analysis}. The seq2seq+pg model is the baseline seq2seq model with pointer-generator; QR-only model is our model variant but trained without coreference modeling.
The overall trend shows that
1) when the dialogue contains coreferences, the joint learning model is more capable of rewriting the query by leveraging its coreference predictions; 2) when coreferences are not present but the query still needs rewriting on account of information omission, the joint model can still perform competitively with the QR-only model.

\subsection{Case Study} \label{sec:case_study}

We demonstrate several examples of query rewrites generated by different models to provide more insights into the task and into the benefits of joint learning. The coreference links predicted by the joint learning model are appended after its generated rewrite. Two examples that require coreference resolution in query rewrite are shown in \tbl{example_coref}. In the left dialogue, \textit{``the song''} in the user query refers back to \textit{``Talking to the Moon''} mentioned in the first user turn. Both seq2seq+pg and QR-only model fail to generate the correct reference in the rewrite, probably because of the high complexity of a long dialogue. The joint learning model not only correctly predicts the coreference link pointing from the mention to its referent in the first turn, but also generates a rewrite perfectly consistent with its coreference prediction. A similar trend can be observed in the right example. The first two models cannot identify which \textit{``Ariana''} to generate, while our model is able to rewrite with the correct one with the aid of the correct coreference resolution.
While our model does well on most of the test cases, there are situations where the joint model fails to predict correctly. A representative failure example is provided in Appendix A.2.

\tbl{example_ellipse} shows an ellipsis example. The implicit location in the user query can be recovered through rewriting by both GPT-2 based models, while the LSTM-based model tends to keep the query.
This indicates that 1) even with the pointer-generator's ability to copy source text, the seq2seq model is not capable enough of handling the difficult information omission rewrite; 2) the joint learning model still performs well on ellipses, while substantially benefiting in coreference cases.

\begin{table}[t]
\centering
\resizebox{\linewidth}{!}{%
\begin{tabu}{l|l}
\tabucline [1pt]{1}
\begin{tabular}[c]{@{}l@{}}Dialogue\\ Context\end{tabular} &
  \begin{tabular}[c]{@{}l@{}}usr: What's the temperature like in Richmond today?\\ sys: The temperature is going to be a warm 85\%,\\\hspace{3.75ex} but there is a chance of rain.\end{tabular} \\ \hline
User Query    & What are the chances of rain today?                              \\ \hline
Rewrite Label & What are the chances of rain today \textbf{in Richmond}? \\ \tabucline [1pt]{1}
seq2seq+pg       & What are the chances of rain today?  \redcross              \\ \hline
QR-only       & What are the chances of rain today \textbf{in Richmond}? \greencheck \\ \hline
Joint         & What are the chances of rain today \textbf{in Richmond}? \greencheck  \\
\tabucline [1pt]{1}
\end{tabu}%
}
\caption{An example with ellipsis from the test set. Rewrites generated by three different models are shown.}
\label{tab:example_ellipse}
\end{table}

\section{Conclusion and Future Work}
We propose a novel joint learning framework for coreference resolution and query rewrite in dialogues.
Modeling coreference resolution not only complements the missing information in query rewrite, but is also beneficial to rewriting anaphoric expressions.
Our joint learning model can predict coreference links between the user query and dialogue context, and generate the rewritten query.
We show that with the aid of coreference resolution, the performance of query rewrite can be substantially boosted.
Furthermore, our model produces competitive results in coreference resolution when compared to state-of-the-art BERT-based systems.
We hope that the presented joint learning task with the release of our query rewrite annotations on the MuDoCo dataset provides a promising research direction in multi-turn dialogue understanding.

One restriction of our model is that by virtue of the model being designed to predict the boundaries of a reference, our model is only able to handle cases involving continuous spans of words.
In addition, the influence of query rewrite on coreference resolution is limited due to the nature of the information flow in our current model design.
Future work will focus on these perspectives.

\section*{Acknowledgements}

The authors would like to thank Hadas Kotek for her help with the data annotation guidelines and the organization of the grading project. The authors would also like to thank Barry-John Theobald, Stephen Pulman, Jason Williams and Murat Akbacak for discussions and feedback, and the anonymous reviewers for their helpful feedback.


\bibliography{anthology,custom}
\bibliographystyle{acl_natbib}

\clearpage
\appendix
\section{Appendices}

\subsection{Training details}
The average run time for training our joint learning model is 6 hours using GTX 1080 Ti. Based on the GPT-2 architecture, our model has 148M parameters. For the attention heads used for predicting the referent in Equation 1, hyper-parameter boundaries for $L'$ and $J'$ are: $1 \leq L' \leq 12$ and $1 \leq J' \leq 12$. The best performance is obtained when only using the last two decode layers ($L'=2$) with 3 attention heads used in each layer ($J'=3$). Hyper-parameters are tuned based on the averaged performance of query rewrite and coreference resolution on the development set.

\subsection{Sample Model-Generated Failure Cases}
We find that our joint model makes mistakes when the coreference signal is ambiguous and there is complex dialogue context (e.g., having multiple person names in an utterance). In a representative example (\tbl{wrong}), even though the joint model predicts one of the coreference links correctly (\textit{one} -> \textit{call}) and generates the corresponding rewritten span (\textit{call from Sana and Erica}), it fails to infer that the pronoun \textit{her} refers to \textit{Deirdre}, and simply ignores the corresponding rewrite. This is likely because there are many female names that the pronoun \textit{her} can refer to in this utterance, and these types of complex cases are too infrequent in the training corpus for the model to learn well.

\subsection{Query Rewrite Annotation Guideline}
The annotation guidelines for collecting query rewrites on the MuDoCo dataset are provided in the following pages. Note that we annotate the rewrite label for every utterance, including the system response, even though they are not used in our experiments.

\begin{table*}[b]
\centering
\resizebox{\linewidth}{!}{%
\begin{tabu}{lllll}
\tabucline [1pt]{1}
\multirow{2}{*}{\begin{tabular}[c]{@{}l@{}}Dialogue\\ Context\end{tabular}} & usr: Answer the call. &  &  &  \\
 & sys: Its Sana and Erica, however, Deirdre on the other line and said its an emergency. &  &  &  \\ \hline
User Query & Very well, cancel the first one and put her through. &  &  &  \\ \hline
Rewrite Label & Very well, cancel the \textbf{call with Sana and Erica} and put \textbf{Deirdre} through. (one -> call, her -> Deirdre) &  &  &  \\ \tabucline [1pt]{1}
seq2seq+pg & Very well, cancel the first \textbf{message} and put her through &  &  &  \\ \hline
QR-only & Very well, cancel \textbf{Deirdre} and put her through. &  &  &  \\ \hline
Joint & Very well, cancel the \textbf{call from Sana and Erica} through. (one -> call, her -> Sana) &  &  & \\
\tabucline [1pt]{1}
\end{tabu}
}
\caption{A complex dialogue example where all systems fail to rewrite correctly. Ground-truth and prediction of coreference are appended correspondingly. Rewritten parts are highlighted in bold.}
\label{tab:wrong}
\end{table*}

\newpage
\clearpage


\section*{Supplementary material}

# MuDoCo Grading Guideline

## Overview

In this project, you will be given a conversation between a user and a virtual assistant. Certain parts of the utterances (both user and assistant) might require previous context to be fully understood. **The goal of the project is to make _minimal_ changes to the current turn so that it can be understood independently, without needing access to the prior context**. This may mean one (or more) of several things:

- Replacing a **pronoun** with its full referent
- Spelling out the content of an **elided phrase**
- Adding elements mentioned in the prior context that are now a part of the understood **Common Ground** of the conversation.

Here is a simple example to get us started. This is a multi-turn conversation that contains a referring expression "that week". If we only had access to the "current" turn, we wouldn't be able to resolve the meaning of "that week"; also, we would incorrectly resolve the intended location of the question to the speaker's location instead of to Sao Paulo, as the previous context makes clear is actually intended.

```
User:      Will next week be a good time to vacation in Sao Paulo?
Assistant: There will be thunderstorms and a lot of rain.
User:      What will the temperature be like that week?            <-- current
```

Our goal in this project is to rewrite **those parts of the current turn that require context**. Here, we can replace "that week" with "next week", and insert "in Sao Paulo" to allow the correct location to be inferred from the request.

```
User:      Will next week be a good time to vacation in Sao Paulo ?
Assistant: There will be thunderstorms and a lot of rain .
User:      What will the temperature be like next week in Sao Paulo?  <-- current
```

**Notice that we do not otherwise alter the rest of the utterance, beyond these minimal changes.**

## Case by Case Guidelines

We introduce our reference / ellipsis resolution guideline in the following categories:

1. **Do rewrite ellipsis as well as references**
2. **Do not paraphrase or summarize but do use the phrasing that appeared in the context**
3. **A special case exception: calls, messages, reminders**
4. **Multiple references**
5. **Data errors**

To formally define what an appropriate "resolution" of a reference / ellipsis is, we introduce the concept of **minimal changes**, which has the following properties we will illustrate with examples below:

1. You may add information that was explicitly uttered in one of the previous turns of the conversation.
2. You should **not** add more information than necessary to make the utterance understandable without additional context, even if such information is available.

3. Special case exception: references to calls, messages, and reminders don't need additional resolution beyond reference to "the call", "the message", or "the reminder".

**<u>Do</u> rewrite ellipsis as well as references**

**Example 1:**

```
User:       Can we expect rain anytime soon?
Assistant:  Rain is possible early next week.
User:       How much chance of rain will there be?          <-- current
```

As we can see, the intended time of the current utterance is understood to be "early next week", since that is the time that is added to the common ground in a previous turn in the conversation as relevant to this weather inquiry. However, if we don't have this information, we would naturally interpret the time of the current turn as "now'". Consequently, we make a minimal change to the current turn by inserting this time information:

```
Desired:    How much chance of rain will there be next week?
Undersired: skip
```

**Example 2:**

```
User:       It's freezing! Is it going to snow?
Assistant:  There's a 65 percent chance of snow tonight
User:       What about this weekend?                         <-- current
```

In this case, the user elided the entire clause "is it going to snow", which we should recover:

```
Desired:    Is it going to snow this weekend?
Undersired: skip
```

**Example 3:**

We may also see cases of noun phrases with some form of ellipsis. These are not implicit arguments but rather explicitly mentioned, but have some *partially* missing information. In such a case we want to fill those out according to our rules to ensure that they are recoverable based on the current turn alone:

```
User:      Play Somewhere Over the Rainbow
Assistant: OK, playing Somewhere Over the Rainbow by Judy Garland.
User:      Can I hear the version by Ariana Grande?
```

The relevant NP here is "the version", which we want to spell out explicitly as "the version of Somewhere Over the Rainbow ":

```
Desired:    Can I hear the version of Somewhere Over the Rainbow by Ariana Grande?
Undersired: Skip
```

## Do not paraphrase or summarize but do use the phrasing that appeared in the context

**Do not rephrase / summarize. Instead, stick to the phrasing that appears in the context as much as possible.** We give several examples of this below.

**Example 1:**

```
User:        What is the latest news on the Brewers ?
Assistant:   There are two recent stories . One is about their lose last night , and th
User:        What happened to him ?
Assistant:   He stepped on his Jesus Aguilar foot .
User:        Then what happened ?
Assistant:   The Brewers bench cleared because they were mad at him .
User:        Can you send more about this to my screen ?      <-- current
```

In this case, the desired behavior is to simply resolve "this" into "this story", rather than providing a paraphrase that collects information from all of the previous turns to create a phrase that wasn't actually uttered in the conversation at all. There is a single story ongoing in this context, so there's not need for a fuller description of this entity:

```
Desired:     Can you send more about this story to my screen ?
Undersired:  Can you send more about The Brewers bench cleared because they were mad at
```

**Example 2:**

```
User:        Can you play Moonlight Sonata by Beethoven?
Assistant:   Yes, you have the live recording of Moonlight Sonata.
User:        I dont want that version. Can you search for the best version on the recor
Assistant:   Yes, the best version is from 1914 and can stream from YouTube to your pla
User:        Okay, stream and save to my playlist under Relaxing.
```

There is an implied argument here (roughly, stream *it* and save *it* ...), which we can recover from the context. Here, we choose the simplest form we can use to complete the utterance, and do not add information across turns.

```
Desired:     Okay, stream the Moonlight Sonata by Beethoven and save it to my playlist
Undersired:  Okay, stream the 1914 version of Moonlight Sonata by Beethoven from YouTub
```

## A special case exception: Calls, messages, reminders

For these entities, you don't need to resolve nominal ellipsis or write out its content in full, just call these entities "the call", "the message", "the reminder". This is an exception to the rules above – there is no need to provide additional information even if it present in the exchange.

**Example 1:**

```
User:        Who is calling me right now?
Assistant:   Marty and Brad are calling you via conference call.
User:        Answer the call but change it to voice only.         <-- current
```

Reference is already made to "the call", and there is only one call referenced in this exchange, so nothing needs to be done here:

```
Desired:    Skip
Undersired: Answer the conference call from Marty and Brad but change it to voice only
```

**Example 2:**

```
User:       I need to send a message.
Assistant:  OK, to who?
User:       To my yoga class.
Assistant:  OK, what do you want to say?
User:       Let them know that class is on for tomorrow.
Assistant:  OK, I have sent your yoga class the message.          <-- current
```

Likewise, there is just one message being discussed here, so there is no need for resolution or added information.

```
Desired:    Skip
Undersired: OK, I have sent your yoga class the message that class is on for tomorrow.
```

**Example 3:**

```
User:       Is there a reminder for this afternoon?
Assistant:  There is one reminder for your sons school.
User:       What does it say?                                      <-- current
```

Here we have a pronoun that needs to be resolved to a noun phrase. The simple noun phrase "the reminder" is all that's needed, please don't add information about the content of the reminder, since this reminder isn't ambiguous with any other reminder in the context.

```
Desired:    What does the reminder say?
Undersired: What does the reminder for my sons school say?
```

## Multiple references

If there are multiple references towards the same entity, you only need to resolve it in the first one. If there is a reference towards an entity in the utterance itself, you don't need to resolve it at all since it's already fully interpretable from that turn alone.

**Example 1:**

```
User:       Send a message to Stephen Bradley.
Assistant:  Okay, what is the message?
User:       Ask him if he wants in on the office lottery pool.     <-- current
```

The current turn contains two pronouns referring to the same individuals. Once the first pronoun is resolved, it provides a reference for the second pronoun within the same sentence. Therefore, we only need to resolve that first

instance (in fact, resolving the second pronoun leads to an unnatural paraphrase that should always be avoided):

```
Desired:    Ask Stephen Bradley if he wants in on the office lottery pool.
Undesired:  Ask Stephen Bradley if Stephen Bradley wants in on the office lottery pool
```

**Example 2:**

```
User:      Set a reminder.
Assistant: Okay, what for?
User:      Remind me to call my daughter for her wedding.         <-- current
```

The pronoun "her" receives a referent within the current turn (namely, "my daughter"), so there is no need to do anything here. You should skip, and not spell out the referent of the pronoun.

```
Desired:    Skip
Undesired:  Remind me to call my daughter for my daughters wedding.
```

### Data errors

Some times the request itself could be faulty, here's a couple of cases::

**Example 1:**
When there is ambiguity in the reference / nominal ellipsis, you should skip

```
User:      Who called me while I was in class?
Assistant: You have a missed call from James and also Sam
User:      Okay, call him back and set my status to online.       <-- current
```

Here since there are two entities, "James" and "Sam", both common male names, it is impossible to determine its referent. You should skip in this case

```
Desired:    Skip
Undesired:  Okay, call James back and set my status to online.
```



\end{document}